\documentclass[11pt]{article}

\usepackage{tikz}
\usepackage{pgf-pie}
\usetikzlibrary{positioning, shapes.geometric, arrows.meta}
\usepackage{subcaption} 
\tikzset{
  block/.style={
  rectangle,
  draw=lightgray,
  minimum width=3cm,
  minimum height=1cm,
  rounded corners,
  text width=3cm,    
  align=center       
  },
  child/.style={
  rectangle, 
  draw=gray, 
  minimum width=3cm, 
  minimum height=1cm,
  rounded corners,
  text width=2.6cm, 
  align=center
  },
  arrow/.style={
  -{Stealth}, 
  thick,
  draw=gray,
  },
}

\usepackage[final]{acl}

\usepackage{times}
\usepackage{latexsym}

\usepackage{multirow} 

\usepackage{subcaption} 

\usepackage{makecell}

\usepackage{algorithm}
\usepackage{algorithmic}

\usepackage[T1]{fontenc}

\usepackage[utf8]{inputenc}

\usepackage{microtype}

\usepackage{inconsolata}

\usepackage{graphicx}

\usepackage{subcaption}

\usepackage{breqn}

\usepackage{xcolor}

\usepackage{enumitem}

\usepackage{amsfonts}

\usepackage{amsmath}

\usepackage{xurl}

\usepackage{tabularx}
\usepackage{booktabs}%
\usepackage{siunitx}
\newcolumntype{R}{>{\centering\arraybackslash}X}

\newcommand{\Ftr}[1]{F_{\mathrm{tr}}(#1)}
\newcommand{\Finf}[2]{F_{\mathrm{inf}}(#1,#2)}
\newcommand{\Accttc}[2]{\mathrm{Acc}^{\mathrm{ttc}}_{#1}(#2)}

\title{FLOP-Efficient Training: Early Stopping Based on Test-Time Compute Awareness}

\author{
  \textbf{Hossam Amer\textsuperscript{1}\thanks{Equal contributions.}},
  \textbf{Maryam Dialameh\textsuperscript{1,2}\footnotemark[1]},
  \textbf{Hossein Rajabzadeh\textsuperscript{1,2}\footnotemark[1]},
  \textbf{Walid Ahmed\textsuperscript{1}},
\\
  \textbf{Weiwei Zhang\textsuperscript{1}},
  \textbf{Yang Liu\textsuperscript{1}},
\\
\\
  \textsuperscript{1}Ascend Team, Huawei Technologies, Toronto, Canada\\
  \textsuperscript{2}University of Waterloo, Waterloo, Canada
\\
  \small{
    \textbf{Correspondence:} \href{mailto:hossam.amer1@huawei.com}{hossam.amer1@huawei.com}
  }
}

\begin{document}
\maketitle
\begin{abstract}
Scaling training compute, measured in FLOPs, has long been shown to improve the accuracy of large language models, yet training remains resource-intensive. Prior work shows that increasing test-time compute (TTC)—for example through iterative sampling—can allow smaller models to rival or surpass much larger ones at lower overall cost. We introduce TTC-aware training, where an intermediate checkpoint and a corresponding TTC configuration can together match or exceed the accuracy of a fully trained model while requiring substantially fewer training FLOPs. Building on this insight, we propose an early stopping algorithm that jointly selects a checkpoint and TTC configuration to minimize training compute without sacrificing accuracy. To make this practical, we develop an efficient TTC evaluation method that avoids exhaustive search, and we formalize a break-even bound that identifies when increased inference compute compensates for reduced training compute. Experiments demonstrate up to 92\% reductions in training FLOPs while maintaining and sometimes remarkably improving accuracy. These results highlight a new perspective for balancing training and inference compute in model development, enabling faster deployment cycles and more frequent model refreshes. Codes will be publicly released.


\end{abstract}

\section{Introduction}

\begin{figure}[t]
    \centering
    \includegraphics[width=\columnwidth]{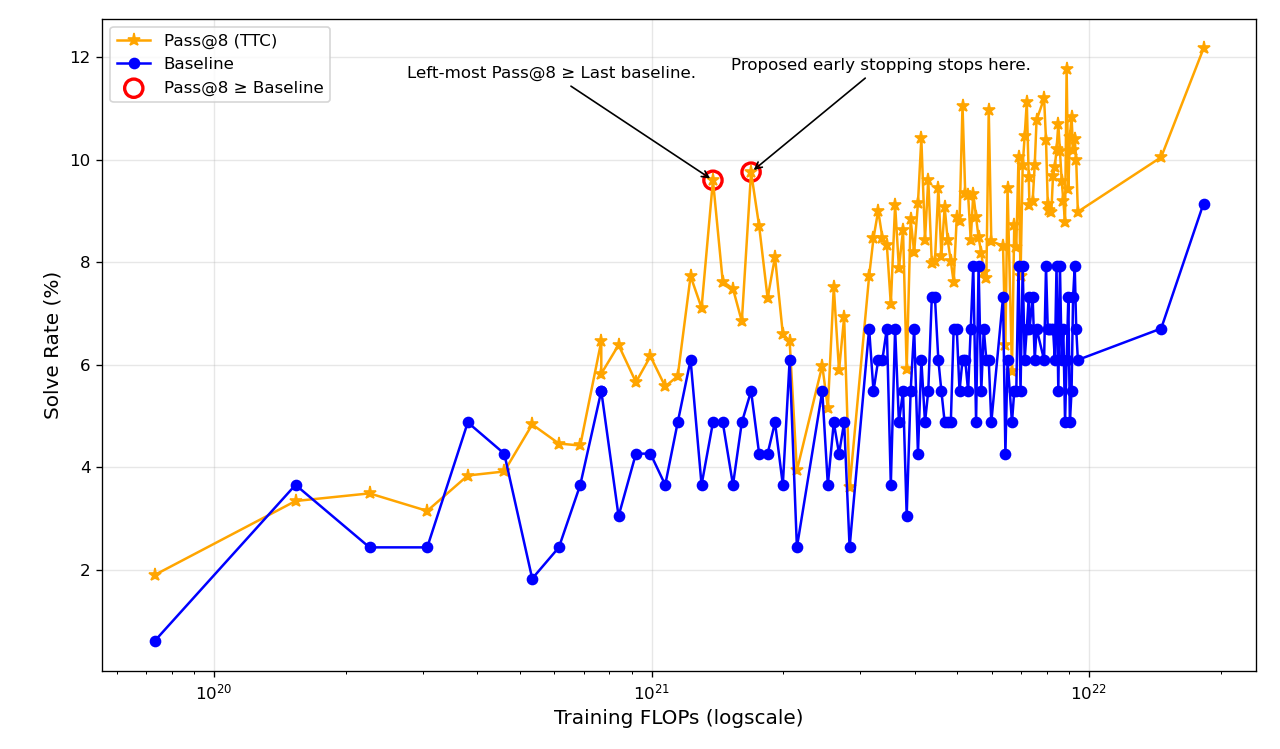}
    \caption{Solve rate (\%) versus training FLOPs for TinyLlama on HumanEval. The solid blue line shows the  baseline checkpoints without TTC, while the solid orange line shows checkpoints using TTC (Pass@8). The red circle indicates that TTC can achieve accuracy at least equivalent to a fully trained checkpoint without TTC, but with significantly fewer training FLOPs. Discussion and results for other $K$ values (e.g., Pass@4) are in Section~\ref{sec:exp1}. Based on our TTC estimation method for K, the predicted value is $K^{*}$ = 8.}
    \label{fig:main-figure}
\end{figure}

Large language models (LLMs) have driven significant progress in natural language processing and beyond \cite{vaswani2017attention, achiam2023gpt, liu2024deepseek}. Open-source architectures such as Llama and Mistral demonstrate strong capabilities in text understanding and generation across diverse tasks. However, these performance gains come at a high cost: training state-of-the-art LLMs demands massive compute, large datasets, and results in considerable financial and environmental burdens \cite{strubell2020energy, hajimolahoseini2023training, tawfilis2025distributed}. For example, training Llama-3.1-405B required 30.84 million GPU hours on NVIDIA H100 hardware, while continued pretraining or finetuning consumes over 2{,}000 H100 GPU hours per billion tokens. Usually, only large labs and companies have access to these resources. These challenges underscore the growing need for different strategies that improve efficiency without compromising accuracy~\cite{meta2024llama31}.

The development of LLMs has been guided by scaling laws describing how accuracy improves with increases in model size, dataset size, and compute. ~\cite{kaplan2020scaling} established that training loss decreases as a power law with respect to model parameters, training tokens, and compute. ~\cite{hoffmann2022training} later proposed the \emph{Chinchilla paradigm}, showing that—under a fixed compute budget—optimal performance is achieved by balancing model size and training tokens, challenging the trend toward excessively large models. 

While these frameworks focus primarily on training compute, recent research has begun to account for inference-time costs \cite{moradi2025continuous}. ~\cite{sardana2023beyond} introduced \emph{inference-aware scaling}, demonstrating that for high-deployment scenarios, smaller models trained longer can yield better overall cost-efficiency. Empirical studies by ~\cite{snell2024scaling} and ~\cite{wu2024inference} show that increasing \emph{test-time compute} (TTC) through methods such as verifier-guided search, or iterative sampling allow smaller models to match or even exceed the performance of much larger models at a lower total compute cost. Related work on repeated sampling and iterative refinement \cite{li2024more, brown2024large, chen2024more} highlights the potential of multipass inference to increase accuracy without training.


Despite these advances, most studies compare fully trained models of different sizes while overlooking training dynamics within a single model, particularly how performance evolves with varying training FLOPs. This gap is critical: if TTC is considered, models could achieve competitive or even superior performance through early stopping. Existing early stopping methods, however, remain TTC-agnostic, focusing only on validation accuracy and neglecting the accuracy gains achievable at inference. Likewise, current TTC techniques, especially those based on repeated sampling, increase inference cost as $K$ grows, where $K$ denotes the number of iterative passes (Pass@$K$).
These techniques do not consider predicting an optimal $K$ to run TTC more efficiently.

These shifting paradigms, from Chinchilla-optimal training to inference-aware scaling, suggest a unifying principle: modern LLM development should jointly optimize training and inference compute. We extend this principle by incorporating \textit{TTC awareness into training}, enabling early stopping decisions based on projected TTC accuracy. Our approach reduces training FLOPs while at least preserving final accuracy, and a new small-$K$ evaluation strategy identifies the optimal $K$ with minimal inference overhead. As Figure~\ref{fig:main-figure} shows, TTC-aware early stopping matches and could sometimes exceed fully trained checkpoint without TTC at a fraction of the compute, \textit{allowing faster model deployments and more frequent model refreshes}.

Overall, this paper contributes the following: 1) Reveals the \emph{TTC-aware training} insight, an insight to reduce total FLOPs while maintaining or improving model accuracy.  
2) Proposes a TTC-aware early stopping algorithm that projects TTC accuracy during training and halts when favorable FLOPs–accuracy trade-offs are achieved. This approach enables faster model refreshes and deployment cycles for large-scale model development, and also benefits academic and open-source labs with limited compute.
3) Develops an efficient TTC estimation method to predict the optimal $K$ from small-$K$ runs, lowering inference cost.  
4) Provides a break-even analysis that identifies when TTC-aware training and inference match or outperform traditional training in total compute.  
5) Empirically validates the effectiveness of TTC-aware training through extensive experiments.

\section{Proposed Method: TTC-Aware Training Procedure}
\label{sec:ttc-aware}



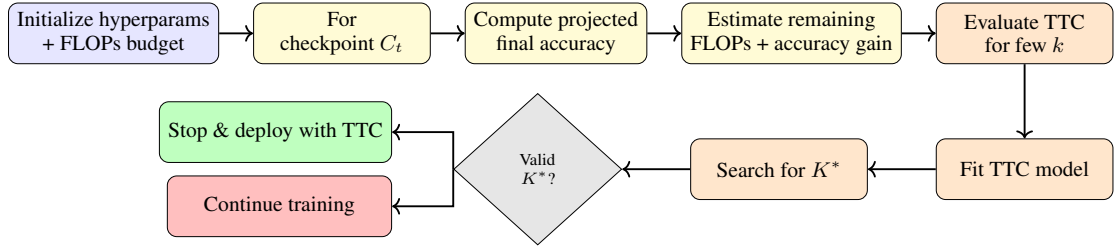
\begin{figure*}[t]
\centering
\resizebox{1.9\columnwidth}{!}{%
\begin{tikzpicture}[
    node distance=1.0cm and 1.0cm,
    every node/.style={font=\footnotesize},
    box/.style={
        rectangle, draw, rounded corners,
        minimum height=0.9cm,
        minimum width=2.6cm,
        align=center
    },
    decision/.style={
        diamond, draw, aspect=1.1,
        align=center,
        inner sep=1.2pt,
        text width=1.8cm,
        font=\scriptsize
    },
    arrow/.style={->, thick}
]

\node[box, fill=blue!10] (init)
    {Initialize hyperparams \\ + FLOPs budget};
\node[box, fill=yellow!20, right=0.5cm of init] (cp)
    {For \\ checkpoint $C_t$};
\node[box, fill=yellow!20, right=0.5cm of cp] (rem)
    {Compute projected \\ final accuracy};
\node[box, fill=yellow!20, right=0.5cm of rem] (estimate)
    {Estimate remaining \\ FLOPs + accuracy gain};
\node[box, fill=orange!20, right=0.5cm of estimate] (evaluate)
    {Evaluate TTC \\ for few $k$};

\node[box, fill=orange!20, below=1.1cm of evaluate] (fit)
    {Fit TTC model};
\node[box, fill=orange!20, left=1.0cm of fit] (search)
    {Search for $K^*$};

\node[decision, fill=gray!20, left=1.0cm of search] (exist)
    {Valid \\ $K^*$?};

\node[box, fill=green!25, minimum width=3.2cm, left=0.9cm of exist, yshift=0.55cm] (deploy)
    {Stop \& deploy with TTC};

\node[box, fill=red!25, minimum width=3.3cm, left=0.9cm of exist, yshift=-0.55cm] (cont)
    {Continue training};

\draw[arrow] (init) -- (cp);
\draw[arrow] (cp) -- (rem);
\draw[arrow] (rem) -- (estimate);
\draw[arrow] (estimate) -- (evaluate);

\draw[arrow] (evaluate) -- (fit);

\draw[arrow] (fit) -- (search);
\draw[arrow] (search) -- (exist);

\draw[arrow] (exist.west) |- node[near start, above]{} (deploy.east);
\draw[arrow] (exist.west) |- node[near start, below]{} (cont.east);

\end{tikzpicture}
}
\caption{Flowchart of the proposed TTC-aware training procedure. The method efficiently selects the intermediate checkpoint with TTC configuration that is superior in terms of FLOPs and accuracy than the fully trained checkpoint, enabling early stopping. Please see Algorithm \ref{alg:ttc_algo} and the associated explanation in Appendix \ref{app:ttc_algo_details}.}
\label{fig:main-method-figure-hor}
\end{figure*}

Figure \ref{fig:main-method-figure-hor} illustrates the overall proposed training procedure with TTC-awareness to achieve FLOP-efficient training. In this procedure, training hyperparameters and training FLOPs budget, $B$, are given before training.  The goal is to train a model such that its final validation accuracy under TTC inference is maximized, subject to the constraint that the total training and inference costs with TTC are less than without TTC. Based on the TTC-awareness insight, we propose a FLOP-efficient training framework that matches or exceeds the validation performance of a fully trained checkpoint under budget $B$. During training, we collect validation accuracy values $\{\mathcal{A}_i\}_{i=1}^n$ from selected intermediate checkpoints $\{t_i\}_{i=1}^n$, and use them to construct a curve-fitted projection of the expected validation accuracy $\hat{\mathcal{A}}(B)$ at budget $B$. As training progresses, this projection is dynamically refined as more checkpoints are observed.

This final accuracy projection relies on empirical observations that typical learning curves follow a saturating pattern: a rapid initial increase in accuracy followed by diminishing returns \cite{kaplan2020scaling}. Thus, we fit the collected validation accuracy data to an exponential saturation function:

\begin{equation}
    \hat{\mathcal{A}}(B) = a \left(1 - e^{-b B} \right) + c
\end{equation}

\noindent where $a$ controls the scale of the gain in accuracy, $b$ determines the rate of convergence (how fast the model saturates), and $c$ represents the base accuracy at initialization. These parameters are learned by minimizing the mean squared error between observed validation accuracies and the fitted curve. The fitted function $\hat{\mathcal{A}}(B)$ is then used to predict final validation performance under budget $B$, guiding early stopping and inference allocation.

During training, validation accuracy typically exhibits fluctuations, which complicates fitting $\hat{\mathcal{A}}(B)$. To mitigate the noisy impact of such fluctuations, we retain only validation points that are monotonically non-decreasing with respect to the best accuracy observed so far; specifically, a point is included if its validation accuracy is greater than or equal to the maximum previously observed value, and discarded otherwise. More generally, this criterion can be relaxed by introducing a user-defined tolerance around the running maximum; however, for simplicity and to avoid introducing additional hyperparameters, we adopt the strict $\geq$ rule in all experiments. We then fit the model exclusively to this filtered, rising portion of the curve. Consequently, the estimator does not introduce additional variance beyond that inherent to the underlying training run. This design choice reflects our objective of accurately estimating the maximum achievable accuracy used to make early stopping decisions (see Figures \ref{fig:exp-curve-fitting-drop} and \ref{fig:exp-curve-fitting}).


To support this framework with efficient inference, we introduce a TTC-aware mechanism that estimates the optimal number of inference samples per query, denoted $K^*$, without incurring excessive compute. Since standard TTC performs inference by sampling the model $K$ times and aggregating the responses, it can become prohibitively expensive if $K$ is large. Instead, we propose to approximate the relationship between validation accuracy under TTC at time $t$ and the number of samples $K$ using a sigmoid curve:

\begin{equation}
    \hat{\mathcal{A}}_K(t) = \frac{L}{1 + e^{-k(K - x_0)}},
\end{equation}

\noindent where $L$ is the maximum achievable TTC accuracy, $k$ controls the growth rate of accuracy with increasing $K$, and $x_0$ is the inflection point, where accuracy reaches half of its maximum. This relation aligns with the intuitive expectation that increasing $K$ will eventually lead to a saturation point, beyond which accuracy gains level off (more evidence is in Figure \ref{fig:ttc-curve-fit} in the appendix).

These parameters are estimated using non-linear least squares fitting over a few sampled values of $K$, obtained by querying the model at small $K$ values (e.g., $K = 1, 2, 4$). Evaluating once at $K$ implicitly yields predictions for all $K' \leq K$, so querying at $K=4$ provides the effective results for $K=1,2,4$ at no additional cost. This formula allows us to extrapolate and predict $K^*$—the smallest value of $K$ that meets both budget and accuracy constraints. The optimal $K^*$ must satisfy the total FLOPs constraint:


{
\begin{align}
\text{F}_{\text{tr}}[t] + \text{F}_{\text{inf}}[t, K^*]
&< \notag\\ \text{F}_{\text{tr}}[B] 
& + \text{F}_{\text{inf}}[B, 1]
\label{eq:flops_inequality}
\end{align}
}

\noindent where $\text{F}_{\text{tr}}[t]$ is the training cost up to timestep $t$, $\text{F}_{\text{inf}}[t, K^*]$ is the cost of evaluating TTC inference at that timestep using $K^*$ samples. In addition, $K^*$ must meet the projected accuracy requirement:

\vspace{-6pt}
\begin{equation}
    \hat{\mathcal{A}}_{K^*}(t) \geq \hat{\mathcal{A}}(B)
\end{equation}
\vspace{-6pt}
\noindent That is, the TTC accuracy at timestep $t$ with $K^*$ samples must be at least as good as the projected final validation accuracy under full budget $B$. If such a $K^*$ exists after a patience window of $p$ checkpoints, training is terminated early; otherwise, training continues to the next checkpoint. 

\begin{algorithm*}
\scriptsize
\caption{TTC-Aware Exponential-Fit Early Stopping}
\label{alg:ttc_algo}
\begin{algorithmic}[1]

\STATE \textbf{Input:} Patience $p$, training budget $B$; \textbf{Init:} $best\_val\_acc, best\_ttc\_val\_acc \gets -\infty$, $patience\_counter \gets 0$
\STATE Define fitting function $f(x)=a(1-e^{-bx})+c$

\FOR{FLOPs $t=3,\!4,\dots$ \textbf{until} $t > B$}

    \STATE Train one step; evaluate to obtain $val\_acc_t$
    \STATE Fit $f(x)$ using observed FLOPs and $\{val\_acc\}$; estimate baseline final accuracy $f(B)$

    \STATE Find \textbf{minimum} $K^*$ satisfying:
\hspace{\algorithmicindent}\parbox[t]{\dimexpr\linewidth-\algorithmicindent\relax}{%
$\begin{aligned}[t]
\Ftr{t}+\Finf{t}{K^*} &< \Ftr{B} + \Finf{B}{1} \quad\& \quad \Accttc{t}{K^*} &\ge f(B)
\end{aligned}$%
}

    \STATE Update $patience\_counter \gets 0$ if $val\_ttc\_acc_t[K^*] > best\_ttc\_val\_acc$

    \STATE $best\_ttc\_val\_acc \gets \max(best\_ttc\_val\_acc, val\_ttc\_acc_t[K^*])$

    \STATE Update $patience\_counter \gets 0$ if $best\_ttc\_val\_acc < f(B)$  else $patience\_counter{+}{+}$
    \IF{$patience\_counter \ge p$}
        \STATE \textbf{Stop}; load best TTC checkpoint; \textbf{break}
    \ENDIF

\ENDFOR

\end{algorithmic}
\end{algorithm*}

\section{Early Stopping Computational Overhead and Break-Even Bound}
\label{sec:break_bound}

The proposed TTC-aware training procedure with early stopping can introduce additional computation, because TTC-driven inference is periodically performed both during training and deployment. It is therefore useful to determine \emph{when} these extra costs are offset by the training savings from TTC-aware early stopping. We refer to this as the \emph{break-even} point: the largest amount of downstream inference that can be performed before the total cost of ``TTC-aware training $+$ deployment'' exceeds the cost of training the same model \emph{without} TTC. 

Here, the total cost explicitly includes both the training cost (which may already incorporate TTC-driven inference) and the deployment cost, i.e., the FLOPs or tokens required to serve end-user queries. Since modern LLMs are not trained once and frozen indefinitely, but rather are \emph{continuously refreshed} through pretraining, finetuning, and ongoing updates \cite{ramaswamy2019federated, gamal2023federated, amer2024device}. Thus, characterizing this bound is especially important: it indicates the consumption of inference tokens before the model should be refreshed.

We denote the number of \emph{training} tokens for the baseline schedule (without TTC) by $N_{\text{train}}$, and the number of \emph{inference} tokens $N_{\text{infer}}$. Here, $N_{\text{infer}}$ accounts for all inference tokens, including those used during TTC sampling as well as tokens served to end users. The ratio of training FLOPs with TTC relative to the baseline is $r \in (0,1]$, where $r < 1$ indicates training savings from earlier stopping. The inference cost of serving the model with TTC is captured by a multiplier $\lambda > 1$, relative to serving the baseline model without TTC. Assuming that training FLOPs per token are roughly $6\times$ the inference FLOPs per token (accounting for forward + backward passes and optimizer updates), the break-even condition — i.e., the maximum number of inference tokens that can be served while ensuring TTC does not exceed the total FLOPs of the baseline without TTC — is:

\begin{equation}
\label{eq:break_even}
N_{\text{infer}} \;\le\; \frac{6 \,(1 - r)}{\lambda - 1} \cdot N_{\text{train}}.
\end{equation}
\noindent A detailed derivation and discussion are provided in Appendix~\ref{sec:break_bound_derivation}.

For illustrative purposes, Table~\ref{tab:bound_table} reports the number of inference samples permitted under our bound, given a model output sequence length of 1024 and a total of 3T training tokens. When $\lambda \approx 1.2\times$, inference with TTC can be applied for up to 70B samples before requiring the next model refresh. This also motivates the importance of the design of even more efficient TTC inference techniques, which we add to future work.

\begin{table}[h!]
\centering
\small 
\begin{tabular}{|c|c|c|c|}
\hline
\textbf{r} & \textbf{$\lambda$} & \textbf{$N_{\text{infer}}$} & \textbf{Inference Samples} \\
\hline
0.4 & 8 & 1.54T & 1.5B \\
0.2 & 16 & 0.96T & 937M \\
0.4 & 1.2 & 54T & 52.7B \\
0.2 & 1.2 & 72T & 70.3B \\
\hline
\end{tabular}
\caption{Illustrative values for $r$, $\lambda$, $N_{\text{infer}}$, and inference samples where each sample has 1024 tokens.}
\label{tab:bound_table}
\end{table}

\section{Experimental Results}
\label{sec:exp1}



\subsection{Experimental Setup}

\noindent \textbf{Models and Checkpoints} We evaluate our approach using three open-source model families: TinyLlama~\cite{zhang2024tinyllama}, Pythia~\cite{biderman2023pythia}, FineMath~\cite{liu2024finemath}, and Qwen3 \cite{yang2025qwen3}. These models were selected because they provide publicly available intermediate checkpoints, unlike most large-scale models where reproducing training would be prohibitively expensive. Table~\ref{tab:models} summarizes the models under evaluation. We use TinyLlama and Pythia to assess the impact of TTC-aware training in a \emph{pretraining} setting, while FineMath enables evaluation in a \emph{continued training} scenario.

\begin{table}[t]
\centering
\scriptsize
\resizebox{\columnwidth}{!}{%
\begin{tabular}{lcccc}
\toprule
 & \textbf{TinyLlama} & \textbf{Pythia} & \textbf{FineMath} & \textbf{Qwen3-A3B-Instruct} \\
\midrule
Size (B) & 1.1 & 1 / 2.8 / 6.9 & 3 & 30 \\
Arch. Base & LLaMA-2 (GQA) & GPT-style & LLaMA-3 & Qwen3 \\
Training Data & \makecell[l]{SlimPajama\\StarCoder} & The Pile
& \makecell[l]{FineWeb-Edu\\FineMath} & Open Data \\
Tokens & $\sim$2T & 300B & 160B & 36T \\
Optimizer & AdamW & Adam + ZeRO & AdamW & -- \\
\bottomrule
\end{tabular}}
\caption{Models used in our experiments.}
\label{tab:models}
\end{table}



\noindent \textbf{Datasets} We evaluate our models on a diverse set of benchmarks spanning code generation, reading comprehension, and mathematical reasoning. Specifically, we include {HumanEval} \cite{li2022evaluating}, a standard benchmark for assessing code synthesis from natural language prompts; {DROP} \cite{dua2019drop}, which tests discrete reasoning and reading comprehension over passages; {Math-500} \cite{hendrycks2021math}, a curated set of advanced mathematics problems; {GSM8k} \cite{cobbe2021training}, a collection of grade-school math word problems designed to evaluate multi-step arithmetic reasoning.

\noindent \textbf{Inference Hardware and Environment} Inference experiments are carried out on a 8xNVIDIA Tesla V100 GPU. Float16 inference is enabled because it speeds up the inference while maintaining accuracy. Our inference environment is
built on top of the InstructEval benchmark \cite{chia2023instructeval}.

\noindent \textbf{TTC Hyperparameter Settings} Our experiments use a temperature of 0.8, a patience value of 10, a maximum input sequence length of 1,024 tokens, and a maximum output length of 512 tokens. The rest of configurations are borrowed from InstructEval repository \cite{chia2023instructeval}.

\subsection{TTC-Aware Training Insight}



\begin{table*}[htbp]
\centering
\small
\resizebox{\textwidth}{!}{%
\begin{tabular}{|c|c|r|r|c|r|r|r|r|r|r|}
\hline
\textbf{Model} & \textbf{Dataset} & \textbf{Step} & \textbf{Tokens} & \textbf{FLOPs} & 
\textbf{Baseline} & \textbf{Pass@4} & \textbf{Pass@8} & \textbf{Pass@16} & \textbf{Pass@32} & \textbf{Pass@64} \\
\hline
Pythia-1B & HumanEval & 80,000  & 167.8T & $9.6\!\times\!10^{20}$ & 1.8  & 2.8  & \textbf{3.8} & 4.9  & 6.4  & 7.9 \\
Pythia-1B & HumanEval & 143,000 & 299.9T & $1.7\!\times\!10^{21}$ & \textbf{3.04} & 4.01 & 5.5  & 7.5  & 9.8  & 12.2 \\
Pythia-1B & GSM8k     & 80,000  & 167.8T & $9.6\!\times\!10^{20}$ & 1.74      & 6.06 & 10.99 & 18.07 & 27.17 & 35.24 \\
Pythia-1B & GSM8k     & 143,000 & 299.9T & $1.7\!\times\!10^{21}$ &  2.20     & 6.97 & 11.22 & 18.34 & 27.52 & 36.16 \\
\hline
Pythia-2.8B & HumanEval & 80,000  & 167.8T & $1.93\!\times\!10^{21}$ & 8.5  & 6.47 & \textbf{8.84} & 11.49 & 14.17 & 17.07 \\
Pythia-2.8B & HumanEval & 143,000 & 299.9T & $3.43\!\times\!10^{21}$ & \textbf{6.7} & 8.38 & 11.02 & 14.01 & 17.53 & 20.73 \\
Pythia-2.8B & GSM8k     & 80,000  & 167.8T & $1.93\!\times\!10^{21}$ & 1.54 & 6.29 & 11.29 & 18.19 & 26.99 & 35.86 \\
Pythia-2.8B & GSM8k     & 143,000 & 299.9T & $3.43\!\times\!10^{21}$ & 2.04 & 6.30 & 11.30 & 18.73 & 27.75 & 38.21 \\
\hline
Pythia-6.9B & HumanEval & 80,000  & 167.8T & $6.7\!\times\!10^{21}$  & 6.8  & 8.17 & 11.09 & 13.96 & 16.98 & 20.12 \\
Pythia-6.9B & HumanEval & 143,000 & 299.9T & $1.2\!\times\!10^{22}$ & 6.8  & 10.28 & 13.22 & 16.40 & 20.15 & 24.39 \\
Pythia-6.9B & GSM8k     & 80,000  & 167.8T & $6.7\!\times\!10^{21}$  & 1.74 & 7.05 & 12.28 & 18.50 & 29.04 & 40.33 \\
Pythia-6.9B & GSM8k     & 143,000 & 299.9T & $1.2\!\times\!10^{22}$ & 2.80 & 7.53 & 13.16 & 19.27 & 29.67 & 40.79 \\
\hline
\end{tabular}
}
\caption{Comparison of intermediate (80k) and final (143k) checkpoints across Pythia models of different sizes on HumanEval and GSM8K. Results are reported using pass@k metrics. \textbf{Bold} entries indicate cases where an intermediate TTC-aware checkpoint matches or surpasses the fully trained baseline at the same or higher $K$, highlighting the consistency and benefit of TTC-aware training.
}
\label{tab:merged_pythia_passk_all}
\end{table*}

TTC-Aware training allows models to reach equal or better accuracy using significantly fewer training FLOPs or less data. This section presents experiments to demonstrate this insight.



\paragraph{TinyLlama on HumanEval and DROP:} As shown in Figure~\ref{fig:main-figure}, an intermediate checkpoint achieves at least the same or even better accuracy than the fully trained baseline without TTC, while providing up to 92.44\% training FLOP savings. In particular, this checkpoint produces a positive accuracy gain of 0.45\%. When accounting for the additional FLOPs consumed by TTC inference during training, the total savings remain substantial at 85.5\%. Although the final checkpoint with Pass@8 attains higher accuracy, this improvement comes at the cost of a substantially larger FLOPs budget compared to the intermediate checkpoint. To further validate this insight, Figure~\ref{fig:drop-tinyllama} exhibits the same trend on DROP, where an earlier checkpoint achieves comparable or better accuracy under a much smaller training budget. This confirms that the TTC-aware training phenomenon is not limited to a single dataset, but rather generalizes across tasks.

\begin{figure}[htpb]
    \centering
    \includegraphics[width=1\linewidth]{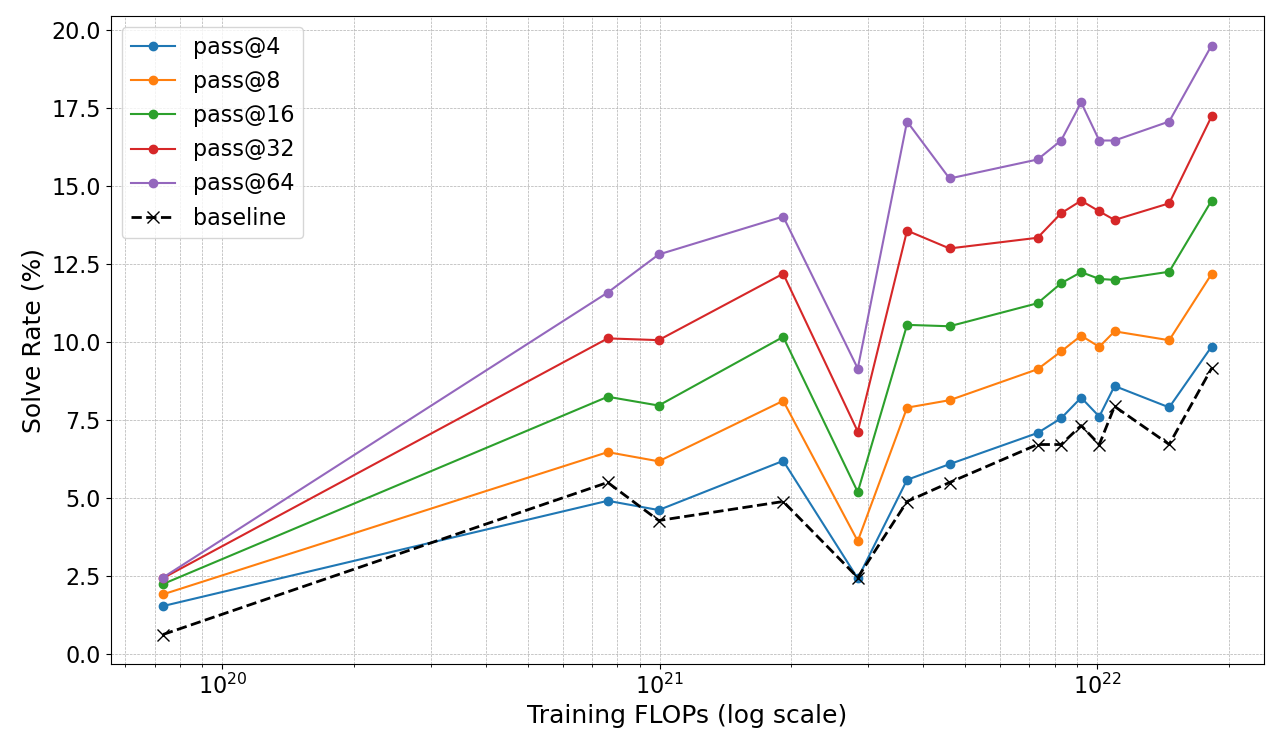}
    \caption{TinyLlama on HumanEval across Pass@$K$. 
Intermediate checkpoints can match or exceed the fully trained baseline with far fewer FLOPs, 
sometimes outperforming the baseline at higher $K$, while the final checkpoint is better only at the cost of much larger FLOPs.}    \label{fig:different_passK_tinyLlama_HumanEval}
\end{figure}

\paragraph{TinyLlama on HumanEval across different Pass@K metrics:} In Figure~\ref{fig:different_passK_tinyLlama_HumanEval}, we observe three consistent trends. First, there always exists an intermediate checkpoint where the Pass@K performance is at least as strong as that of the fully trained baseline without TTC, while requiring substantially fewer FLOPs. Second, when comparing across different $K$ values, if the intermediate checkpoint uses a \emph{larger} $K$ than the fully trained model (e.g., Pass@8 or Pass@16 vs. Pass@4), it can often maintain or even surpass the accuracy of the fully trained baseline while still providing FLOPs savings. Under TTC, if the first checkpoint used a smaller 
$K$ than the fully trained, it attains lower accuracy. To reach comparable performance, training must proceed to the full budget. Third, when both models are compared at the same $K$, the fully trained baseline attains higher accuracy, but only at the cost of a substantially larger FLOPs budget.






\paragraph{Pythia on HumanEval, GSM8K:} For each Pythia 1B/2.8B/6.9B model, we compare an intermediate checkpoint at 80k training steps with the fully trained baseline at 143k training steps (see Table~\ref{tab:merged_pythia_passk_all}). Across both HumanEval and GSM8k, we observe the same TTC-aware trends identified earlier with TinyLlama. Intermediate checkpoints with TTC match or exceed the performance of fully trained baselines while requiring substantially fewer FLOPs. For example, on HumanEval, the 1B Pythia model at 80k achieves Pass@8 accuracy (3.8\%) comparable to its 143k baseline without TTC (3.04\%), while the 2.8B model at 80k with Pass@8 (8.84\%) not only surpasses its 143k baseline (6.7\%) but also outperforms the much larger 6.9B baseline at 143k (6.8\%).

\noindent A similar pattern appears on GSM8k: the 2.8B checkpoint at 80k matches or exceeds the 143k baseline across all Pass@K values, and the 6.9B checkpoint at 80k with TTC outperforms its fully trained counterpart without TTC. Furthermore, when comparing different $K$ values within the same model size, if the intermediate checkpoint uses a larger $K$ than the fully trained baseline, its accuracy is often higher. For example, Pythia-2.8B at 80k with Pass@8 (11.29\%) surpasses Pythia-2.8B at 143k with Pass@4 (6.3\%). Consistent patterns are also observed across Pythia model sizes, reinforcing the findings from TinyLlama: the TTC-aware effect generalizes across model scales and datasets, demonstrating that intermediate checkpoints can achieve comparable or better performance at a fraction of the training cost.


\paragraph{FineMath with actual TTC techniques:} In all previous experiments, we focused on Pass@$K$, which captures the potential performance of a model under multiple sampled outputs. To extend this to more advanced TTC techniques beyond Pass@$K$, we evaluate several approaches: (1) the Baseline without TTC, (2) Pass@$K$, (3) Majority Voting, (4) Diverse Tree Search (DVTS) with naive weighting, (5) DVTS with weighted aggregation, and (6) Compute-Optimal Search, which selects the best-performing method among Majority Voting and the two DVTS variants~\cite{beeching2024scalingtesttimecompute, snell2024scaling}.

\noindent In DVTS with weighted aggregation, identical responses are grouped, and the final answer is chosen based on the highest cumulative reward across all identical responses. In contrast, DVTS with naive weighting treats each generated response independently, selecting the one with the highest reward model (RM) score as the final output. 

\noindent We apply these techniques to the FineMath model in a continued-training setup at 10B, 30B, 40B, and 50B training tokens within the mathematics domain and evaluate on the Math-500 benchmark. 


\noindent An important observation from Table~\ref{tab:finemath-results} is that even when a lower training budget underperforms at a given Pass@$K$, there often exists a larger $K$ where it surpasses higher-budget checkpoints evaluated at smaller $K$. For example, at $K{=}2$ the 10B checkpoint underperforms the 50B one (15.05\% vs.\ 18.28\%), but at $K{=}4$, the 10B checkpoint outperforms the 50B checkpoint with $K{=}2$ (24.73\% vs.\ 18.28\%). This pattern recurs across multiple settings (e.g., 10B vs.\ 40B), illustrating the broader TTC-aware insight in a continued training scenario as shown with FineMath.


We further find that this TTC-aware insight generalizes beyond Pass@$K$ to verifier-based and aggregation metrics. For instance, the 10B checkpoint with $K{=}16$ under the \textsc{DVTS-Weighted} verifier achieves a 19.35\% solve rate, surpassing the 50B checkpoint without TTC (15.05\%). Similarly, the 50B checkpoint improves from 13.98\% at $K{=}2$ to 21.51\% at $K{=}4$, demonstrating the consistent performance gains achieved with larger $K$. To reach comparable or better accuracy than the 50B checkpoint, one could instead use the 10B checkpoint with $K{=}32$ (33.33\%) or the 30B checkpoint with $K{=}16$ (24.73\%) or the 40B checkpoint with $K{=}32$ (41.94\%). Across all TTC techniques evaluated, we observe that solve rates generally increase as $K$ grows. However, verifier-based TTC scores are typically lower than raw Pass@$K$ values, underscoring the importance of designing strong verifiers for realistic evaluation. Among the TTC techniques, \textsc{DVTS-Weighted} consistently yields the best overall performance, while the Compute-Optimal Search provides an approximate upper bound over the tested methods. These findings reinforce that the TTC-aware insight persists even under verifier-based and aggregation metrics, further validating the generality of the TTC framework across both metric types and model scales.

\paragraph{TinyLlama on Math-500}
\label{sec:tinyllama_math_500}

We evaluate TinyLlama on the Math-500 benchmark, a standard dataset for mathematical reasoning tasks. Math-500 is particularly challenging for TinyLlama because the model was not originally trained on sufficient math-domain data, resulting in a 0\% score for the original fully trained checkpoint. To address this, we conducted additional training to include enough math data, despite our limited compute resources. Table~\ref{hyperparameters-tinyllama} details the full training configuration.  

\begin{figure}[htbp]
    \centering
    \includegraphics[width=1.1\linewidth]{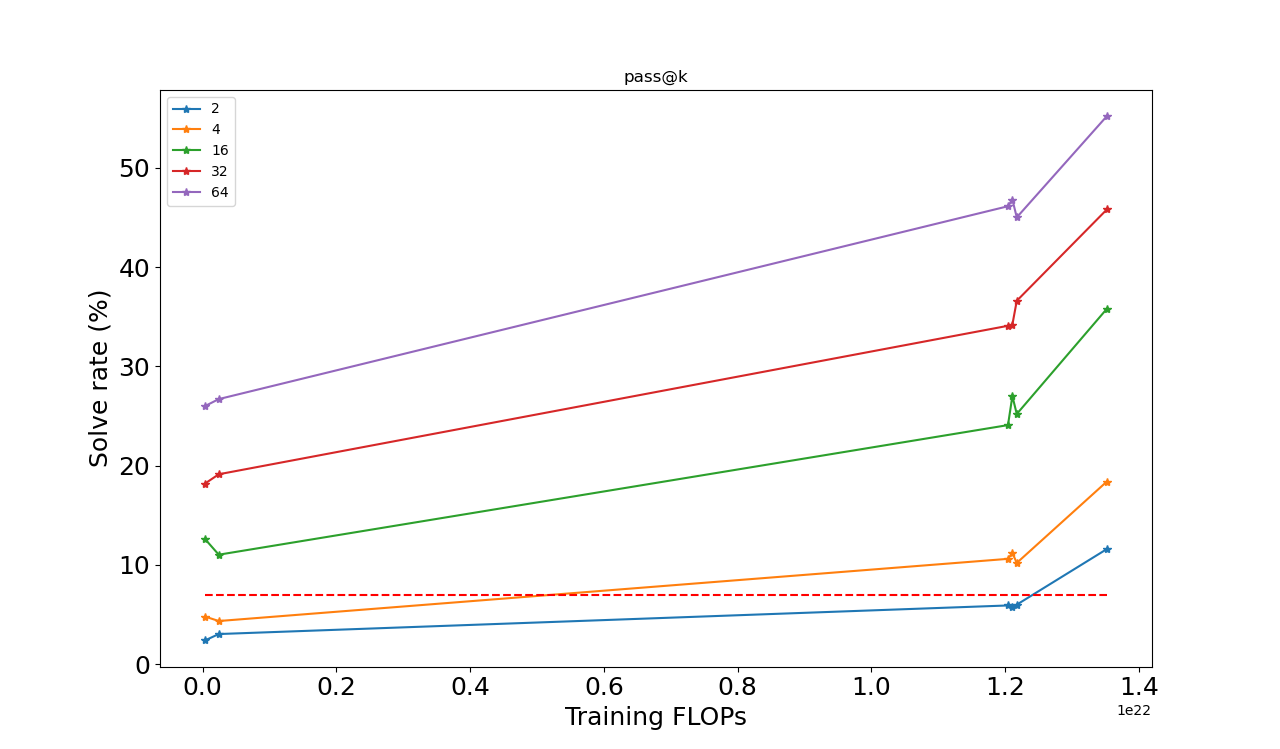}
    \caption{TinyLlama on Math-500 dataset with different Pass@K. The dotted horizontal line represents the score of the fully trained TinyLlama without TTC. Challenging setup given the TinyLlama available checkpoints accuracy on Math-500 and our compute resources.}
    \label{fig:tllama_math500_passK}
\end{figure}

Due to these compute constraints, we were only able to use a subset of intermediate checkpoints. Nevertheless, even with this partial coverage, the baseline without TTC shows a roughly monotonic increase across available checkpoints, albeit with some variation in scores. As shown in Figure~\ref{fig:tllama_math500_passK}, the results reveal two clusters of points: one corresponding to checkpoints trained from scratch on cosmopedia dataset with varying amounts of FLOPs appearing on the left, and another corresponding to fine-tuned checkpoints with varying amount of FLOPs appearing on the right.  

Despite these practical limitations, the TTC-aware insight remains visible. This demonstrates both the promise of TTC for improving training efficiency and the challenges in fully exploiting it across models and checkpoints, highlighting the need for appropriate training resources and setup in future work.


\begin{figure}[htpb]
    \centering
\includegraphics[width=1\linewidth]{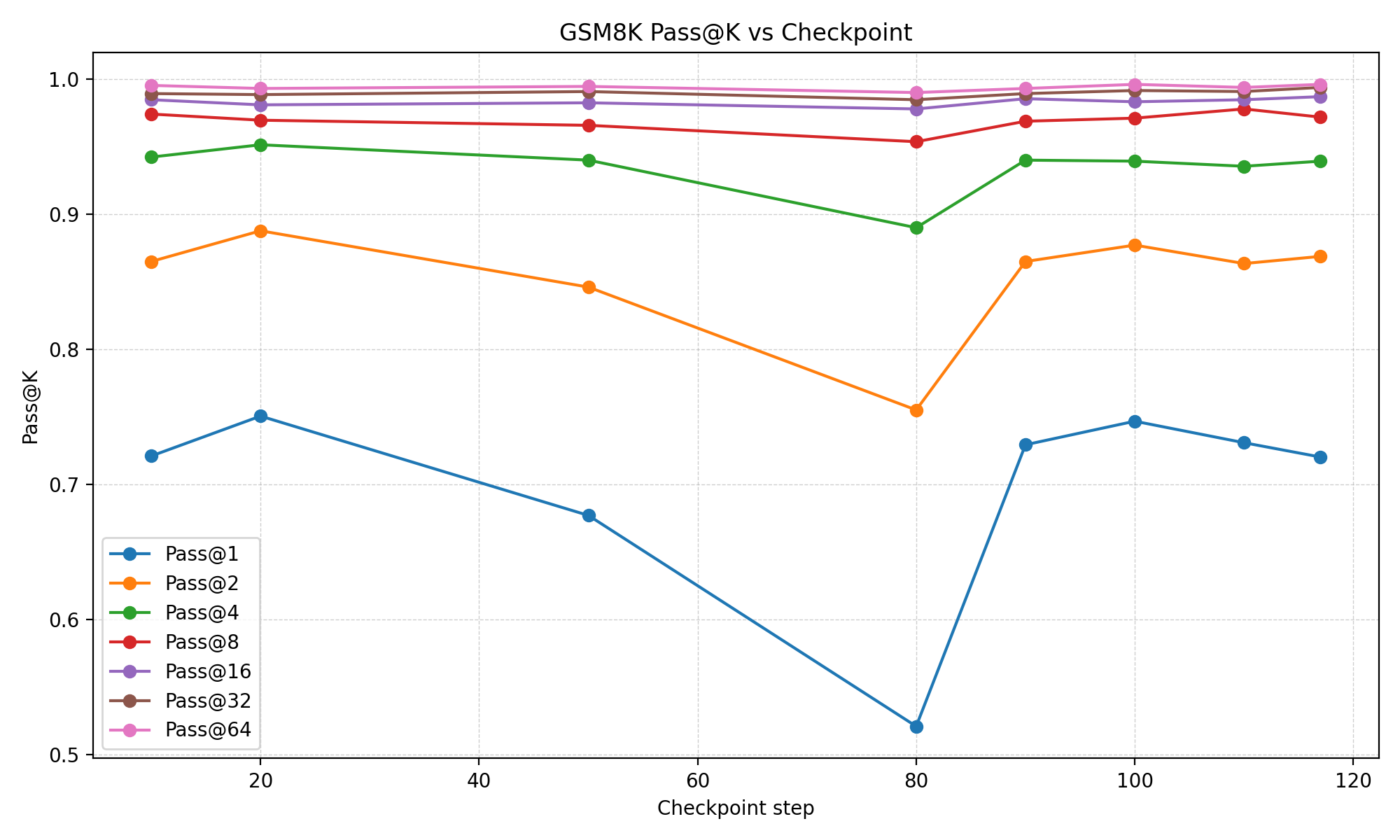}
    \caption{Qwen3-30B-A3B-Instruct on GSM8K (1 epoch). Pass@K over intermediate checkpoints shows non-monotonic behavior; TTC selects an intermediate checkpoint that outperforms the final checkpoint.}    \label{fig:qwen3-30b}
\end{figure}

\paragraph{Results on Large Frontier Models:} 
While evaluating our method on frontier-scale LLMs is limited to the availability of intermediate checkpoints, we created the following setting: we fine-tuned Qwen3-30B-A3B-Instruct for 1 epoch on GSM8K and saved intermediate checkpoints throughout training and applied our TTC selection over them. As shown in Figure \ref{fig:qwen3-30b}, even in this large-model setting, we observe the same TTC-aware effect: the early TTC-selected intermediate checkpoint with Pass@2 can outperform the final checkpoint by more than 10\%, demonstrating that TTC remains beneficial beyond the small-model regime.

\begin{table}[t]
\centering
\resizebox{\columnwidth}{!}{%
\setlength{\tabcolsep}{4pt}
\begin{tabular}{c|c|c|c|c|c|c|c}
\toprule
\textbf{Train Tokens} & \textbf{$K$} & \textbf{Baseline} & \textbf{Pass@K} & \textbf{Majority} & \textbf{Naive} & \textbf{Weighted} & \textbf{CTS} \\
\midrule
\multirow{6}{*}{10B}
 & 2  & 12.90 & 15.05 & 12.90 & 12.90 & 12.90 & 13.98 \\
 & 4  & 12.90 & 24.73 & 11.83 & 11.83 & 12.90 & 16.13 \\
 & 16 & 12.90 & 53.76 & 19.35 & 21.51 & 19.35 & 29.03 \\
 & 32 & 12.90 & 67.74 & 26.88 & 23.66 & 33.33 & 44.09 \\
 & 64 & 12.90 & 74.19 & 25.81 & 24.73 & 34.41 & 39.78 \\
\midrule
\multirow{6}{*}{30B}
 & 2  & 7.53 & 12.90 & 8.60 & 11.83 & 11.83 & 12.90 \\
 & 4  & 7.53 & 21.51 & 11.83 & 19.35 & 20.43 & 21.51 \\
 & 16 & 7.53 & 46.24 & 19.35 & 24.73 & 24.73 & 33.33 \\
 & 32 & 7.53 & 61.29 & 23.66 & 23.66 & 27.96 & 36.56 \\
 & 64 & 7.53 & 68.82 & 31.18 & 26.88 & 35.48 & 45.16 \\
\midrule
\multirow{6}{*}{40B}
 & 2  & 11.83 & 13.98 & 11.83 & 12.90 & 12.90 & 13.98 \\
 & 4  & 11.83 & 24.73 & 15.05 & 18.28 & 17.20 & 21.51 \\
 & 16 & 11.83 & 51.61 & 29.03 & 22.58 & 29.03 & 38.71 \\
 & 32 & 11.83 & 67.74 & 37.63 & 27.96 & 41.94 & 50.54 \\
 & 64 & 11.83 & 73.12 & 39.78 & 32.26 & 49.46 & 52.69 \\
\midrule
\multirow{6}{*}{50B}
 & 2  & 15.05 & 18.28 & 13.98 & 13.98 & 13.98 & 15.05 \\
 & 4  & 15.05 & 29.03 & 18.28 & 18.28 & 21.51 & 24.73 \\
 & 16 & 15.05 & 59.14 & 36.56 & 27.96 & 37.63 & 45.16 \\
 & 32 & 15.05 & 66.67 & 38.71 & 32.26 & 39.78 & 49.46 \\
 & 64 & 15.05 & 76.34 & 40.86 & 34.41 & 44.09 & 53.76 \\
\bottomrule
\end{tabular}
}
\caption{Solve Rate (\%) of \textit{FineMath} checkpoints (10B, 30B, 40B, 50B tokens) across different $K$. The Verifier is Skywork-o1-Open-PRM-Qwen-2.5-1.5B. We report Baseline without TTC, Pass@$k$, Majority, DVTS Naive, DVTS Weighted, and Compute-Optimal Search (CTS) aggregation methods on Math-500.}
\label{tab:finemath-results}
\end{table}

\begin{figure}[!t]
    \centering
    \includegraphics[width=\columnwidth]{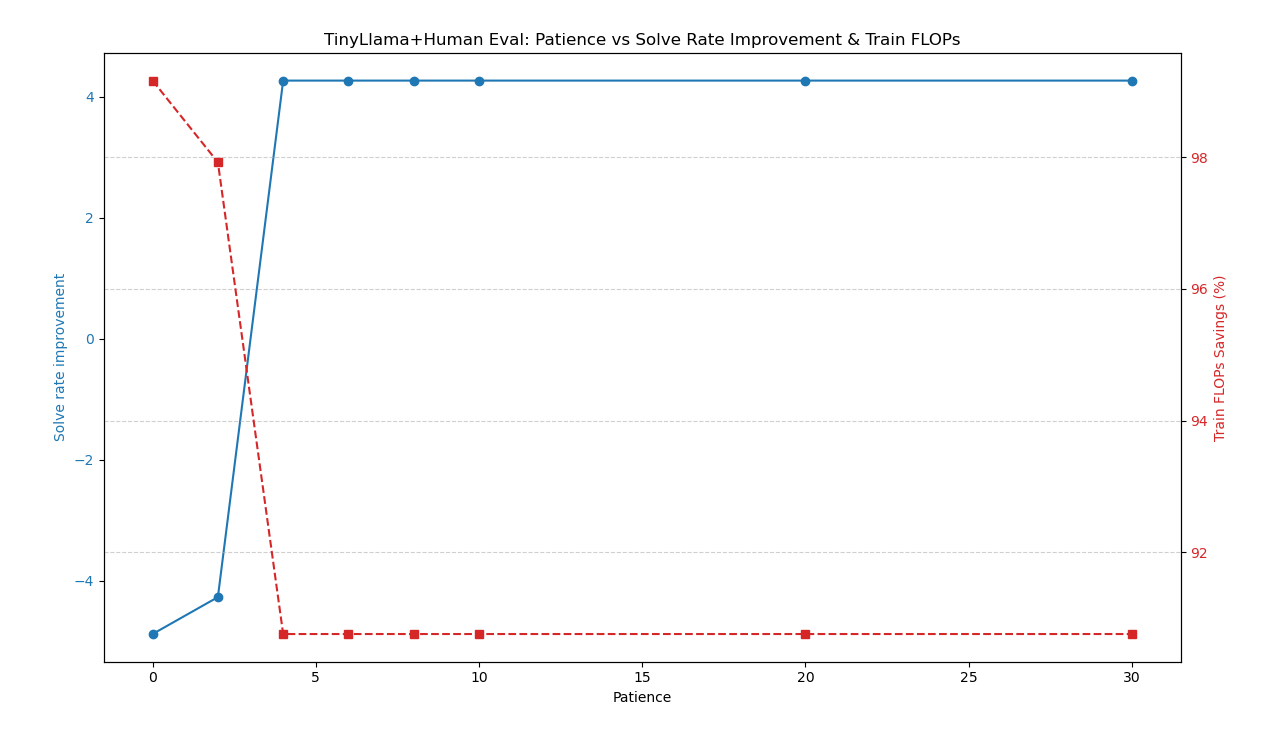}
    \caption{Effect of patience on TinyLlama (HumanEval). Higher patience improves TTC Pass@16 but increases training FLOPs; a mid-range value ($\approx10$) offers the best trade-off, while overly large patience reduces early-stopping benefits.}

    \label{fig:patience-ablation}
\end{figure}
\vspace{-16pt}

\subsection{Early Stopping Analysis and Results}

The proposed early stopping algorithm achieves noticeable gains as given in Table~\ref{tab:early-stopping-datasets}. In TinyLlama with the HumanEval dataset, it reaches Pass@8 accuracy roughly 0.6\% higher than the fully trained checkpoint without TTC while providing 90.7\% training FLOP savings, and delivers a 4.3\% improvement in Pass@16 with 90.76\% training FLOP savings. Using our early stopping method, Pass@8 reduces training FLOPs by 90.7\% relative to the baseline, while we could practically reach 92.44\% overall FLOP savings. Similar trends hold across datasets and model families, including Pythia~\cite{biderman2023pythia}. Overall, TTC-aware training can save up to 92\% of training FLOPs without sacrificing accuracy and often improves it.

To further benchmark the proposed early stopping algorithm, we implement PyTorch-style naive early stopping and evaluate it on TinyLlama–HumanEval~\cite{ignite-earlystopping-doc}. We vary the patience from 0 to 100 with $\epsilon=0$. We report $\boldsymbol{\Delta}$, the accuracy difference between the early-stopped model and the fully trained baseline; FLOPs saving, the fraction of training compute saved; and $\boldsymbol{\Delta_{\text{TTC}}}$, the accuracy difference after applying TTC Pass@8 to the early-stopped checkpoint. As shown in Table~\ref{tab:naive-earlystop-patience-ablation}, naive early stopping exhibits a poor trade-off: it either halts very early, yielding large FLOPs savings but substantial accuracy degradation, or stops near full training with negligible savings. Applying TTC post hoc does not meaningfully improve this trade-off. Overall, naive early stopping does not perform as the proposed method, which can achieve up to 92\% FLOPs savings while improving accuracy by 0.44\%.

\begin{table}[htbp]
\centering
\caption{Effect of validation fluctuation filtering on early stopping for TinyLlama.}
\label{tab:early-stopping-datasets}
\resizebox{\columnwidth}{!}{%
\begin{tabular}{lcccc}
\toprule
& \multicolumn{2}{c}{\textbf{HumanEval}} 
& \multicolumn{2}{c}{\textbf{DROP}} \\
\cmidrule(lr){2-3} \cmidrule(lr){4-5}
\textbf{Method}
& \textbf{FLOPs Saving}
& \textbf{Acc. Gain}
& \textbf{FLOPs Saving}
& \textbf{Acc. Gain} \\
\midrule
Early stopping w/ fluctuations 
& 56.8\%
& +2.05\%
& 72.7\%
& +2.08\% \\

Early stopping w/o fluctuations
& 90.7\%
& +0.6\%
& 87.0\%
& +1.0\% \\
\bottomrule
\end{tabular}
}
\end{table}

\begin{table}[htbp]
\centering
\scriptsize
\caption{Relative accuracy and computational cost (FLOPs) of naive early stopping versus the fully trained TinyLlama checkpoint on HumanEval.}
\label{tab:naive-earlystop-patience-ablation}
\setlength{\tabcolsep}{3pt}      
\renewcommand{\arraystretch}{1.05}
\begin{tabular}{c
  S[table-format=3.2]
  S[table-format=2.2]
  S[table-format=+1.2]}
\toprule
\textbf{Patience} &
{$\boldsymbol{\Delta}$ (\%)} &
{\textbf{FLOPs Saving} (\%)} &
{$\boldsymbol{\Delta_{\text{TTC}}}$ (\%)} \\
\midrule
0   & -4.27 & 97.92 & -5.31 \\
6   & -3.66 & 95.80 & -2.69 \\
8   & -3.05 & 93.28 & -1.42 \\
10  & -3.05 & 93.28 & -1.42 \\
20  & -1.22 & 70.24 & +0.19 \\
50  &  0.00 &  0.00 & +3.03 \\
100 &  0.00 &  0.00 & +3.03 \\
\bottomrule
\end{tabular}
\end{table}

\section{Conclusion}
\label{sec:conclusion}
We introduced \emph{TTC-aware training}, which selects an intermediate checkpoint together with a minimal test-time compute setting $K^*$ to maximize quality per total FLOPs, rather than defaulting to the final checkpoint at budget $B$. Across multiple model scales and settings, this yields substantial savings with no quality loss: we achieve up to \textbf{92.44\%} training-FLOPs reduction while preserving (and often improving) downstream performance. For example, an \textbf{80k-step} Pythia-2.8B checkpoint with TTC reaches \textbf{8.84\%} Pass@8 on HumanEval versus \textbf{6.7\%} for the fully trained baseline, and verifier-based TTC on Math-500 improves from \textbf{15.05\%} (50B-token baseline) to \textbf{19.35\%} at a \textbf{10B-token} checkpoint with $K{=}16$. Our TTC-aware early stopping strategy further strengthens this result: on TinyLlama--HumanEval, it improves Pass@8 by roughly \textbf{+0.6\%} over the fully trained checkpoint (without TTC) while saving \textbf{90.7\%} training FLOPs, and achieves \textbf{+4.3\%} Pass@16 with \textbf{90.76\%} FLOPs savings. Finally, our break-even bound shows that these training savings can fund TTC for \textbf{$\sim$70B} downstream samples (for $\lambda\!\approx\!1.2\times$), making TTC-aware training especially attractive for frequently refreshed models.

\section*{Limitations}
\noindent \textbf{Models and Checkpoints.}
Our study is constrained to publicly available checkpoints up to 6.9B parameters. Larger models such as Llama-3.x-405B do not release intermediate checkpoints, preventing direct evaluation.

\noindent \textbf{Datasets.}
Some model–dataset combinations (e.g., TinyLlama on Math-500) perform poorly due to limited domain pretraining and compute, restricting evaluation breadth.

\noindent \textbf{TTC Inference Efficiency.}
Realizing TTC gains in deployment may require efficient inference strategies (e.g., beam search, diverse search) to control latency and cost. Integrating such optimizations remains an important direction.

\noindent \textbf{Verifier-Based Evaluation.}
We primarily evaluate verifier-based methods in continued pretraining. A broader study across multiple verifiers and architectures is valuable future work, but our focus here is demonstrating FLOP reductions while maintaining accuracy.

\noindent \textbf{Scaling Laws.}
A full TTC-aware training scaling law—characterizing the optimal stopping point—remains open and is an important avenue for future research.

\bibliography{custom}
\newpage
\appendix




\section{Derivation: Break-even Condition for Test-Time Compute (TTC) with Continuous Refreshing}
\label{sec:break_bound_derivation}

\noindent
Define the following quantities per unit time (e.g., per year):
\begin{itemize}
  \item $\mathrm{FLOPs}_{\text{train}}$: training FLOPs for the baseline (no TTC).
  \item $\mathrm{FLOPs}_{\text{infer}}$: inference FLOPs for the baseline (no TTC).
  \item $r\in(0,1]$: ratio of training FLOPs under TTC to baseline training FLOPs, i.e. $r=\dfrac{\mathrm{FLOPs}_{\text{train}}^{\text{TTC}}}{\mathrm{FLOPs}_{\text{train}}}$.
  \item $\lambda_{\text{TTC}}>0$: inference cost multiplier when using TTC (per-token inference cost relative to the ``no TTC'' baseline).
  \item $\lambda_{\text{base}}\ge 1$: inference cost multiplier already present in the baseline deployment (equals $1$ when the baseline has no extra inference overhead).
  \item $f>0$: refresh frequency (number of training+deployment cycles per unit time). For example, if the model is retrained and deployed three times per year, then f =3. For the algebra below $f$ cancels and therefore does not affect the bound.
\end{itemize}

\noindent
Total FLOPs per unit time for the baseline and for TTC-aware operation are
\begin{align}
C_{\text{baseline}}
&= f\cdot\big(\mathrm{FLOPs}_{\text{train}} + \lambda_{\text{base}}\,\mathrm{FLOPs}_{\text{infer}}\big),\\
C_{\text{TTC}}
&= f\cdot\big(r\,\mathrm{FLOPs}_{\text{train}} + \lambda_{\text{TTC}}\,\mathrm{FLOPs}_{\text{infer}}\big).
\end{align}

\noindent Here we assume that we can train and/or deploy the underlying model with TTC controlled by $\lambda_{\text{base}}$ or $\lambda_{\text{TTC}}$.

\noindent
Requiring TTC to be no more expensive than the baseline gives
\[
C_{\text{TTC}} \le C_{\text{baseline}}.
\]

Cancel \( f \) (assuming \( f > 0 \)):
\begin{equation*}
\begin{split}
r\,\mathrm{FLOPs}_{\text{train}} 
&+ \lambda_{\text{TTC}}\,\mathrm{FLOPs}_{\text{infer}} \\
&\le \mathrm{FLOPs}_{\text{train}} 
+ \lambda_{\text{base}}\mathrm{FLOPs}_{\text{infer}}
\end{split}
\end{equation*}

\noindent
Thus the FLOPs-level break-even condition is
\[
\boxed{(1-r)\,\mathrm{FLOPs}_{\text{train}} \;\ge\; (\lambda_{\text{TTC}} - \lambda_{\text{base}})\,\mathrm{FLOPs}_{\text{infer}}.}
\]

\noindent
Next express the bound in terms of token counts. Let $N_{\text{train}}$ be the number of training tokens per refresh (baseline) and $N_{\text{infer}}$ the number of inference tokens served before the next refresh. Using the common approximation that one training token costs roughly $6\times$ the FLOPs of one inference token (accounting for forward + backward + optimizer)~\cite{kaplan2020scaling, anil2023palm},
\[
\frac{\mathrm{FLOPs}_{\text{train}}}{\mathrm{FLOPs}_{\text{infer}}}
\approx \frac{6\,N_{\text{train}}}{N_{\text{infer}}}.
\]
Substitute this ratio into the FLOPs inequality:
\begin{equation*}
\begin{aligned}
(1-r)\cdot \frac{6\,N_{\text{train}}}{N_{\text{infer}}}\cdot \mathrm{FLOPs}_{\text{infer}} \\
\!\!\! \ge (\lambda_{\text{TTC}} - \lambda_{\text{base}})\,\mathrm{FLOPs}_{\text{infer}}
\end{aligned}
\end{equation*}

Cancel $\mathrm{FLOPs}_{\text{infer}}>0$ and solve for $N_{\text{infer}}$:
\[
(1-r)\cdot 6\,N_{\text{train}} \;\ge\; (\lambda_{\text{TTC}} - \lambda_{\text{base}})\,N_{\text{infer}}
\]
\[
{N_{\text{infer}} \;\le\; \dfrac{6\,(1-r)}{\lambda_{\text{TTC}} - \lambda_{\text{base}}}\;N_{\text{train}}.}
\]

Setting $\lambda_{\text{base}}=1$ recovers the special case where the baseline has no extra inference overhead:
\[
\boxed{
N_{\text{infer}} \leq \frac{6(1 - r)}{(\lambda - 1)} \cdot N_{\text{train}}
}
\]


\noindent
\textbf{Remarks:}
\begin{itemize}
  \item The denominator $\lambda_{\text{TTC}} - \lambda_{\text{base}}$ must be positive for the bound to be meaningful. If $\lambda_{\text{TTC}}\le\lambda_{\text{base}}$, the accuracy of the intermediate checkpoint with TTC will not be better than the fully trained checkpoint.
  \item For $\lambda_{\text{base}}>1$, the denominator $(\lambda_{\text{TTC}} - \lambda_{\text{base}})$ decreases, 
    which increases the upper bound on $N_{\text{infer}}$. 
    This means that as the baseline incurs more inference overhead, TTC-aware training allows proportionally more inference tokens before the break-even point is reached. For this reason, we set $\lambda_{\text{base}}$ to 1 to minimize the overhead of the baseline.
  \item This bound gives the maximum number of inference tokens that can be served between refreshes while keeping the total FLOPs (training plus deployment) at or below the non-TTC baseline.
\end{itemize}

\section{Curve Fitting Results and Analysis}

\begin{figure*}[htbp]
    \centering



    \begin{subfigure}[t]{\textwidth}
        \centering
        \includegraphics[width=0.48\linewidth]{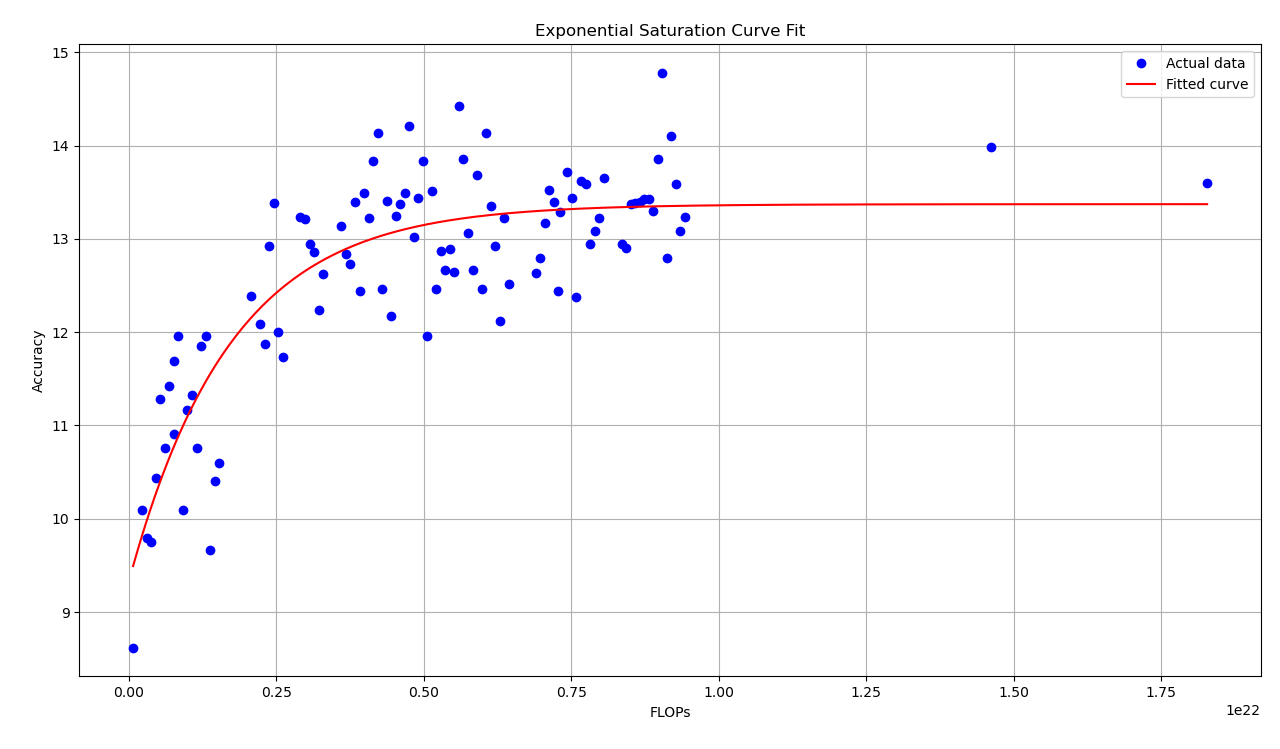}
        \caption{Curve fit across all checkpoints with fluctuations.}
        \label{fig:exp-fit-all-drop}
    \end{subfigure}

    \vspace{0.5em} 

    \begin{subfigure}[t]{\textwidth}
        \centering
        \includegraphics[width=0.48\linewidth]{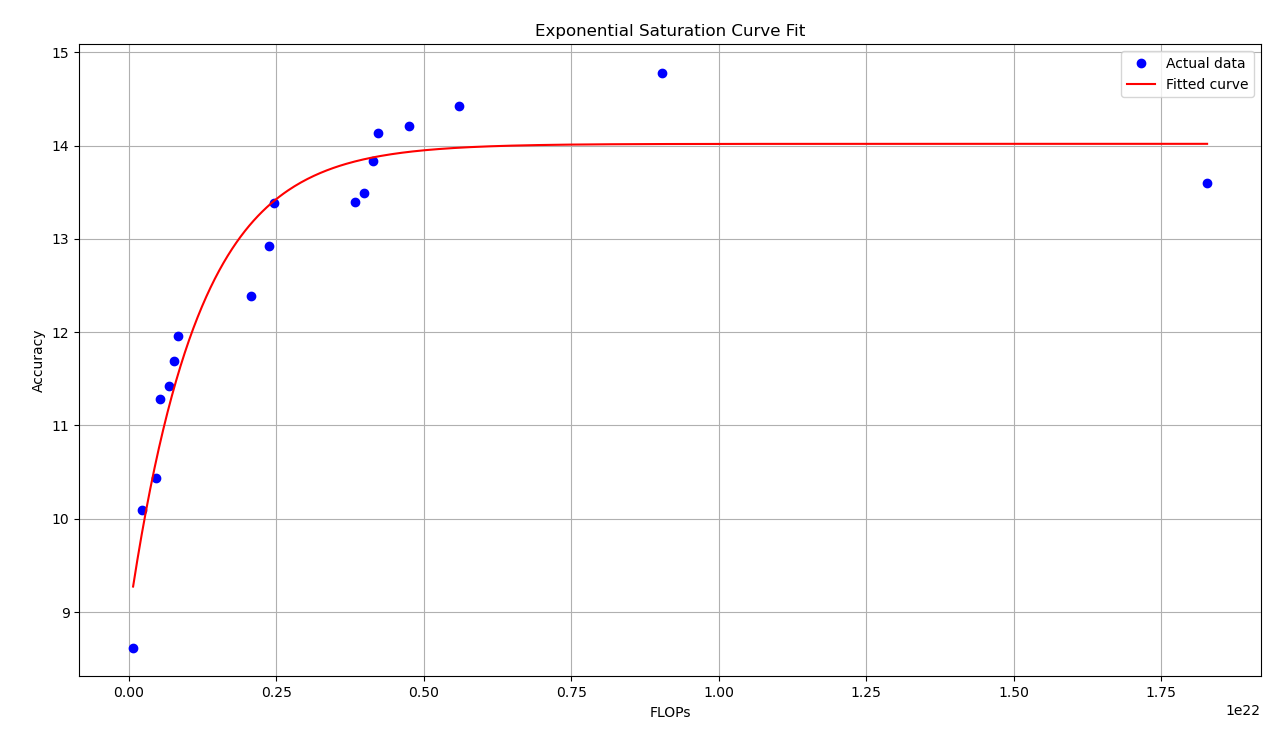}
        \caption{Curve fit across all checkpoints without fluctuations.}
        \label{fig:exp-fit-all-nooutliers-drop}
    \end{subfigure}

    \caption{Exponential saturation curve fitting results for \textbf{TinyLlama on DROP}.
    (a) shows the fit when using all available checkpoints. (b) shows the fit when using all available checkpoints and ignoring fluctuations.
    In all cases, the red curve represents the fitted function and the blue points denote actual data, where each point corresponds to a TinyLlama checkpoint evaluated on the DROP dataset. After fluctuation filtering, the relative error between the fitted checkpoint and the final observed checkpoint is as low as +0.4\% }
    \label{fig:exp-curve-fitting-drop}
\end{figure*}

Figures~\ref{fig:exp-curve-fitting-drop} and ~\ref{fig:exp-curve-fitting} illustrate exponential curve fitting on the checkpoints, showing minimal fitting error. Because we exclude non-monotonic points in curve fitting, we remove the expected fluctuations in the validation accuracy. Shown in Table \ref{tab:early-stopping-datasets}, we observe that removing these fluctuations result in better FLOPs savings -- from 56.8\% to 90.7\% -- in TinyLlama on HumanEval. Thus, our curve fitting procedure for training is more stable and has more FLOPs savings while at least maintaining the accuracy. Additionally, Figure~\ref{fig:ttc-curve-fit} presents sigmoid curve projections, which closely track TTC accuracy across checkpoints at different FLOP levels.

To further guard against potential fitting failures, we incorporate \emph{patience}, a standard hyperparameter in early stopping. Figure~\ref{fig:patience-ablation} illustrates the impact of varying patience on the solve rate improvement for TTC Pass@16—relative to the fully trained checkpoint—and the corresponding training FLOPs savings for TinyLlama on the HumanEval dataset. A clear tradeoff emerges: increasing patience generally improves the solve rate but also decreases the fraction of training FLOPs saved. While larger patience values yield slightly higher improvements, the FLOPs savings decrease, resulting in diminishing returns in computational efficiency. An intermediate patience value of around 10 appears optimal, providing substantial solve rate gains while keeping training FLOPs within a reasonable range. This underscores the importance of carefully tuning patience to balance performance and efficiency.



\noindent To further verify robustness, we also repeated the same experiments at temperatures 0.7, 0.8, and 0.9, each with three seeds. The resulting average standard deviation across checkpoints remains close to 0.1, confirming stability across decoding settings as well.




\section{TTC Latency Overhead}
A common concern is that test-time compute (TTC) increases deployment latency because decoding is repeated $K$ times (e.g., Pass@$K$, majority vote, or verifier-based search). In our framework, this overhead is explicitly controlled in two ways. First, the early-stopping procedure estimates the \emph{smallest} TTC budget $K^*$ that meets the target accuracy while still satisfying the total-compute constraint in Eq.~\ref{eq:flops_inequality}. Empirically, $K^*$ is modest: on both TinyLlama/HumanEval and Qwen3-30B/GSM8k, intermediate checkpoints can already match or exceed the fully trained baseline with small $K$ (Figure~\ref{fig:different_passK_tinyLlama_HumanEval}), and on FineMath/Math-500, a \textbf{10B-token} checkpoint with verifier-based DVTS at \textbf{$K{=}4$} reaches \textbf{24.73\%}, outperforming a \textbf{50B-token} checkpoint with \textbf{$K{=}2$} (\textbf{18.28\%}) (Table~\ref{tab:finemath-results}). These results indicate that very large $K$ is not necessary for deployment; TTC gains are realizable at low-to-moderate $K$.

Second, we quantify the \emph{end-to-end} overhead through the break-even bound in Eq.~\ref{eq:break_even}, which guarantees when ``TTC-aware training + deployment'' remains cheaper than training and deploying the fully trained baseline. This bound uses $\lambda$ (the inference-cost multiplier induced by TTC) and $r$ (the training-FLOPs ratio after early stopping). Table~\ref{tab:bound_table} shows that even with non-trivial TTC overhead, the training savings can fund a large amount of downstream inference before the next refresh: for example, with a modest overhead $\lambda{\approx}1.2\times$, the bound permits \textbf{54T--72T} inference tokens, corresponding to \textbf{52.7B--70.3B} samples of length 1024 tokens. Even for larger overheads (e.g., $\lambda{=}8$ or $\lambda{=}16$), the break-even point remains on the order of \textbf{0.96T--1.54T} inference tokens (Table~\ref{tab:bound_table}). 

Finally, the practical overhead during training is negligible: curve fitting (exponential/sigmoid over a few points) is closed-form and run only when patience triggers a check, while TTC sampling is \emph{fully accounted for} by the break-even analysis. Overall, TTC-aware early stopping trades a small, controlled inference-time multiplier for a large reduction in training compute, enabling faster model development and more frequent refresh cycles. Economically, even in our \textbf{1.1B} TinyLlama experiment, TTC-aware early stopping reduces the effective training from \textbf{1,657 GPU-days} to at least \textbf{127 GPU-days} (Pass@8), roughly cutting cost from \textbf{135k} to \textbf{10k} on Azure V100 pricing; since training cost scales approximately linearly with model size, the relative savings become even more meaningful at \textbf{70B--400B} scale.

\section{Fitting for the Final Accuracy and TTC}

\begin{figure*}[htbp]
    \centering



    \begin{subfigure}[t]{\textwidth}
        \centering
        \includegraphics[width=0.48\linewidth]{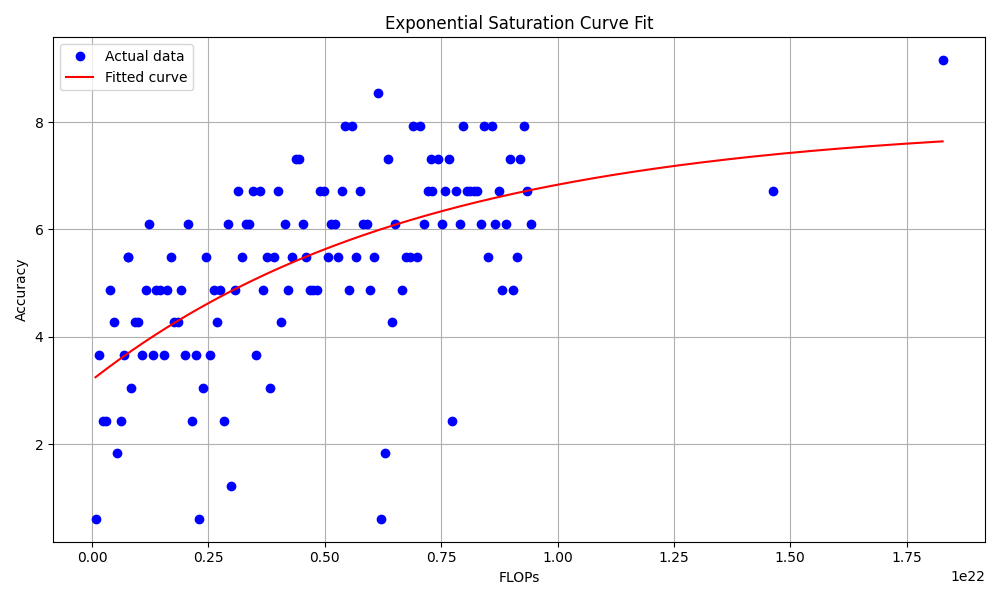}
        \caption{Curve fit across all checkpoints with fluctuations.}
        \label{fig:exp-fit-all}
    \end{subfigure}

    \vspace{0.5em} 

    \begin{subfigure}[t]{\textwidth}
        \centering
        \includegraphics[width=0.48\linewidth]{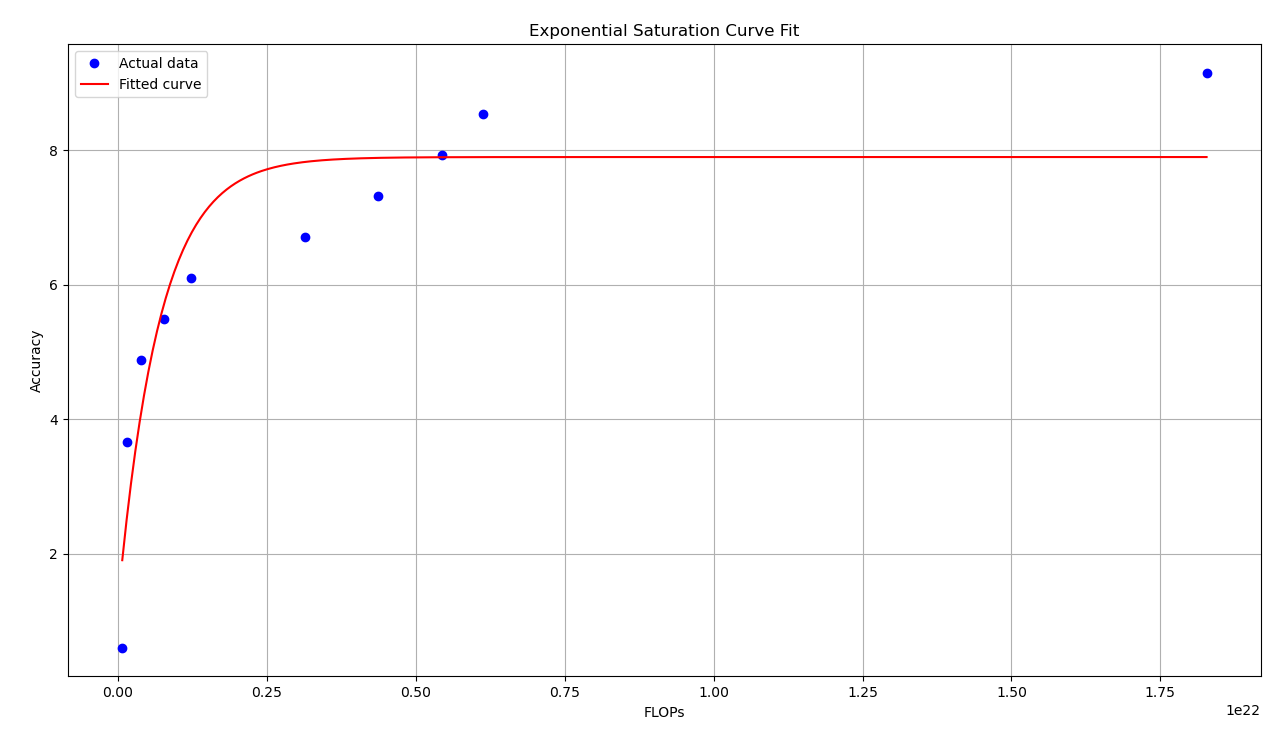}
        \caption{Curve fit across all checkpoints without fluctuations.}
        \label{fig:exp-fit-all-nooutliers}
    \end{subfigure}

    \caption{Exponential saturation curve fitting results.
    (a) shows the fit when using all available checkpoints for \textbf{TinyLlama on HumanEval}. (b) shows the fit when using all available checkpoints and ignoring fluctuations.
    In all cases, the red curve represents the fitted function and the blue points denote actual data, where each point corresponds to a TinyLlama checkpoint evaluated on the HumanEval dataset. After fluctuations filtering, relative error between the fitted checkpoint and the last actual checkpoint = -1.8\%}
    \label{fig:exp-curve-fitting}
\end{figure*}

\begin{figure*}
    \centering
    \includegraphics[width=\textwidth]{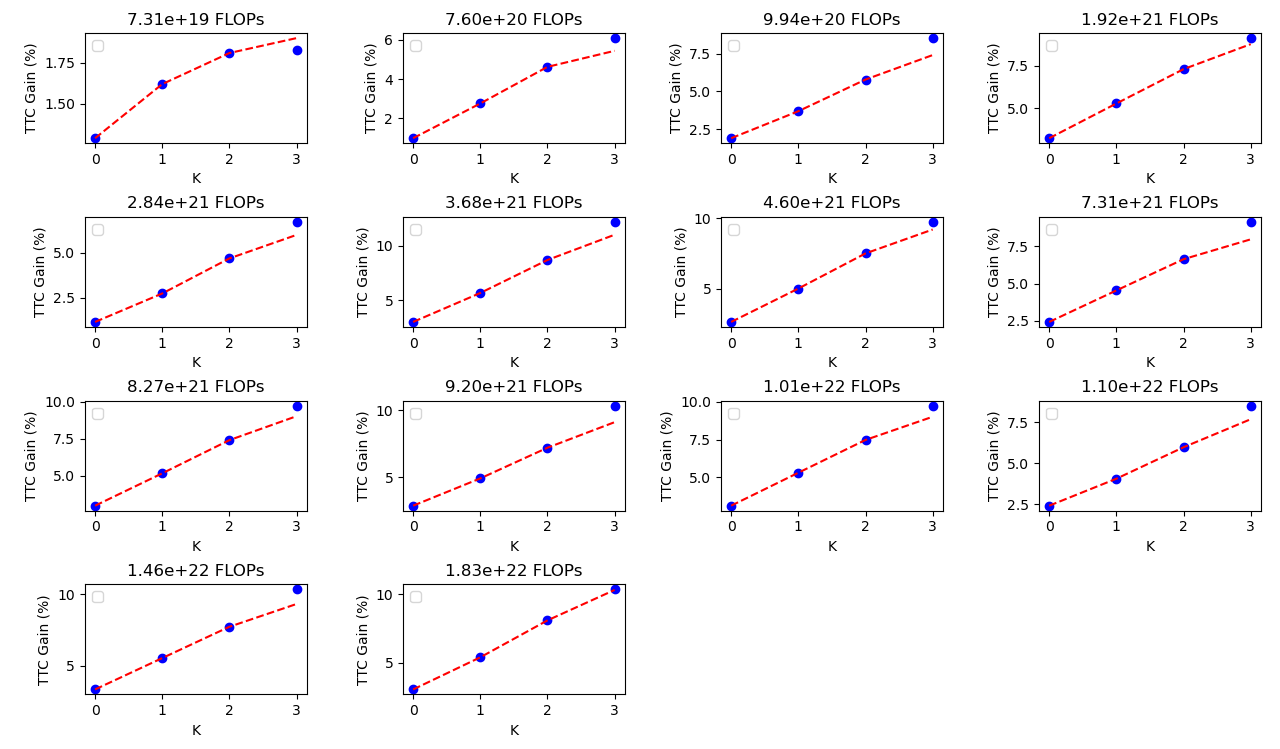}
    \caption{Sigmoid fitting for the parameter $K$ in TTC at different FLOPs. TTC gain is recorded for $K = 1, 2, 4$ and the fitted curve is used to estimate accuracy at $K = 8$ on TinyLlama-1B with the HumanEval dataset.}
    \label{fig:ttc-curve-fit}
\end{figure*}

Figures \ref{fig:exp-curve-fitting} and \ref{fig:ttc-curve-fit} show the successful fitting of for the exponential saturating function and sigmoid functions, respectively. 


\begin{figure*}[htbp]
    \centering
    \includegraphics[width=0.63\textwidth]{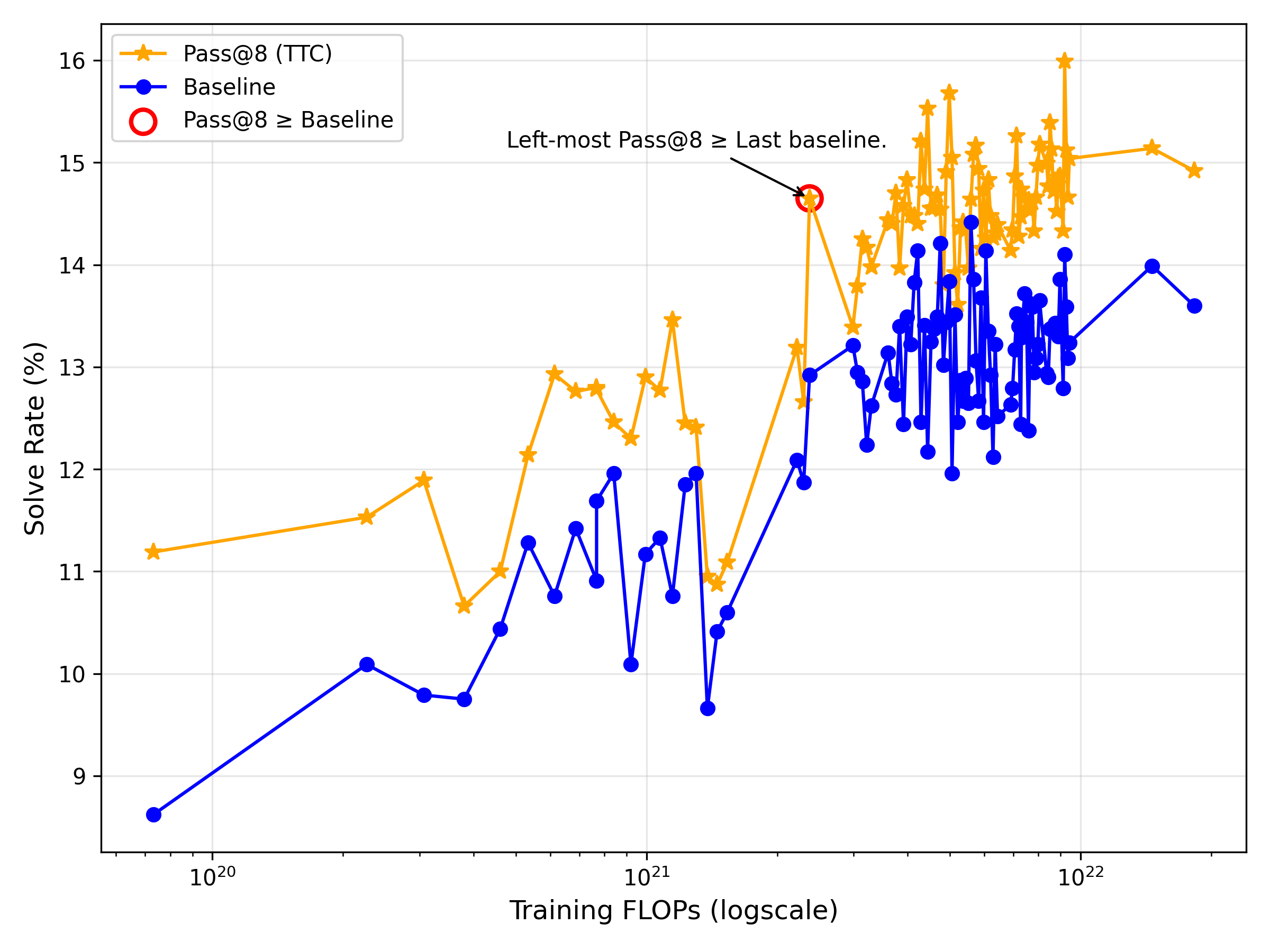}
    \caption{Results of TTC-aware training on the DROP dataset using TinyLlama-1B. Similar to Figure~\ref{fig:main-figure}, applying TTC allows early stopping while saving up to 92\% of the training FLOPs without sacrificing final accuracy. This demonstrates that TTC awareness generalizes across datasets and is not limited to a single benchmark.}
    \label{fig:drop-tinyllama}
\end{figure*}

\section{Details of TTC-Aware Exponential-Fit Early Stopping}
\label{app:ttc_algo_details}

Algorithm~\ref{alg:ttc_algo} implements a deployment-aligned early-stopping rule that decides when to stop training and how much test-time compute to use at the selected checkpoint. Instead of optimizing for the final checkpoint at the full training budget $B$, the algorithm searches for an intermediate checkpoint at FLOPs $t$ paired with a minimal TTC budget $K^*$ that (i) meets or exceeds the predicted fully-trained accuracy and (ii) is cheaper in total compute than training to $B$ and deploying with standard decoding.

\paragraph{Notation and objective.}
At each training time $t$, we consider two costs: training FLOPs $\Ftr{t}$ and inference FLOPs under TTC $\Finf{t}{K}$, where $K$ denotes the number of samples/decoding attempts used by the TTC method (e.g., Pass@$K$, majority vote, DVTS-based search).
We also denote the TTC-validated accuracy at time $t$ using budget $K$ by $\Accttc{t}{K}$, which may be estimated on a held-out validation set.
The core goal is to find a \emph{compute-dominating} TTC checkpoint, i.e.,
\begin{equation}
\begin{aligned}
\Ftr{t}+\Finf{t}{K} &< \Ftr{B}+\Finf{B}{1}, \\ and \\
\Accttc{t}{K} &\ge Acc(B).
\end{aligned}
\end{equation}

\noindent where $Acc(B)$ is the accuracy of the (unknown) fully trained checkpoint.
Since $Acc(B)$ is not available before completing training, we estimate it via learning-curve fitting.

\paragraph{Step 1: Exponential learning-curve fit to forecast $f(B)$.}
Line 2 defines an exponential saturation model
$f(x)=a(1-e^{-bx})+c$,
where $x$ is the cumulative training FLOPs (or steps mapped to FLOPs).
At each iteration $t$ (lines 4--6), the algorithm fits $f(\cdot)$ to the observed validation accuracies $\{val\_acc\}$ collected so far and uses the fitted curve to forecast the baseline final accuracy at the full budget, $f(B)$.
This forecast acts as a surrogate for the fully trained checkpoint performance and is updated online as new checkpoints are observed.

\paragraph{Step 2: Selecting the minimal TTC budget $K^*$.}
Given the current checkpoint at $t$, the algorithm chooses the \emph{smallest} TTC budget $K^*$ that satisfies two constraints (line 8):
\begin{align}
\Ftr{t}+\Finf{t}{K^*} &< \Ftr{B} + \Finf{B}{1},
\label{eq:appendix_compute_constraint}
\\
\Accttc{t}{K^*} &\ge f(B).
\label{eq:appendix_accuracy_constraint}
\end{align}
The compute constraint in Eq.~\eqref{eq:appendix_compute_constraint} enforces that ``stop at $t$ + deploy with TTC'' is strictly cheaper than ``train to $B$ + deploy with standard decoding.''
The accuracy constraint in Eq.~\eqref{eq:appendix_accuracy_constraint} enforces that TTC at checkpoint $t$ meets (or exceeds) the predicted fully trained accuracy.
By minimizing $K^*$, the algorithm avoids unnecessarily large TTC overhead and directly targets the smallest deployment-time cost that achieves the desired quality.

\paragraph{Step 3: Tracking the best TTC checkpoint.}
Once $K^*$ is determined, the algorithm evaluates (or looks up) TTC validation accuracy at that $K^*$, denoted $val\_ttc\_acc_t[K^*]$.
If this value improves upon the best TTC validation accuracy seen so far, $best\_ttc\_val\_acc$, we reset the patience counter (line 10) and update $best\_ttc\_val\_acc$ (line 12).
This ensures we retain the best \emph{deployable} checkpoint under the TTC budget dictated by the inequality and target accuracy.

\paragraph{Step 4: Patience-based stopping criterion.}
The stopping condition (lines 14--17) triggers when the algorithm has observed no TTC-improving checkpoint for $p$ consecutive checks \emph{after} the TTC performance has reached the forecasted baseline.
Concretely, if $best\_ttc\_val\_acc \ge f(B)$, the patience counter increments; otherwise it is reset (line 14).
When $patience\_counter \ge p$, training stops and the algorithm returns the best TTC checkpoint found so far (line 16).
This design prevents premature stopping before TTC catches up to the predicted final baseline, while still terminating soon after the training trajectory saturates in a deployment-relevant sense.

\paragraph{Practical notes.}
(i) The fit is performed on a small set of observed points and updated online; in practice, it is invoked only when a patience check is performed.
(ii) $\Finf{t}{K}$ can be measured directly (e.g., wall-clock or FLOPs) or modeled as approximately linear in $K$ for sampling-based TTC, which is sufficient for selecting $K^*$.
(iii) The algorithm is agnostic to the particular TTC procedure: $\Accttc{t}{K}$ may come from Pass@$K$, majority vote, verifier-guided search, or other TTC policies, as long as accuracy increases monotonically (or approximately monotonically) with $K$.

\begin{table*}[htbp]
\centering
\small
\caption{Pretraining configuration for TinyLLaMA in LLaMA-Factory \cite{zheng2024llamafactory}.}
\begin{tabular}{|l|l|}
\hline
\textbf{Category} & \textbf{Configuration} \\
\hline
Model & \begin{tabular}[c]{@{}l@{}}
\texttt{quantization\_method}: None \\
\texttt{train\_from\_scratch}: True \\
\texttt{disable\_gradient\_checkpointing}: False \\
\end{tabular} \\
\hline
Method & \begin{tabular}[c]{@{}l@{}}
\texttt{stage}: pt \\
\texttt{do\_train}: True \\
\texttt{finetuning\_type}: full \\
\texttt{deepspeed}: examples/deepspeed/ds\_z3\_config.json \\
\end{tabular} \\
\hline
Dataset & \begin{tabular}[c]{@{}l@{}}
\texttt{dataset}: cosmopedia-v2, From HuggingFaceTB/smollm-corpus \\
\texttt{cutoff\_len}: 2048 \\
\texttt{preprocessing\_num\_workers}: 120 \\
\texttt{streaming}: False, \texttt{packing}: True \\
\texttt{overwrite\_cache}: True \\
\texttt{dataloader\_num\_workers}: 8 \\
\end{tabular} \\
\hline
Training & \begin{tabular}[c]{@{}l@{}}
\texttt{per\_device\_train\_batch\_size}: 16 \\
\texttt{gradient\_accumulation\_steps}: 4 \\
\texttt{learning\_rate}: 2.0e-5 \\
\texttt{lr\_scheduler\_type}: cosine \\
\texttt{warmup\_ratio}: 0.005 \\
\texttt{fp16}: True, \texttt{bf16}: False \\
\texttt{ddp\_timeout}: 180000000 \\
\texttt{num\_train\_epochs}: 1 \\
\end{tabular} \\
\hline
Evaluation & \begin{tabular}[c]{@{}l@{}}
\texttt{val\_size}: 5000 \\
\texttt{per\_device\_eval\_batch\_size}: 1 \\
\texttt{eval\_strategy}: steps \\
\texttt{eval\_steps}: 200 \\
\end{tabular} \\
\hline
Output & \begin{tabular}[c]{@{}l@{}}

\texttt{logging\_steps}: 1, \texttt{save\_steps}: 200 \\
\texttt{plot\_loss}: True \\
\texttt{overwrite\_output\_dir}: True \\
\texttt{save\_total\_limit}: 500 \\
\texttt{report\_to}: tensorboard \\
\end{tabular} \\
\hline
\end{tabular}
\label{hyperparameters-tinyllama}
\end{table*}

\end{document}